\documentclass[letterpaper]{article} 
\usepackage{aaai25}  
\usepackage{multirow,makecell,booktabs,hhline,amsmath,amsfonts,arydshln}
\usepackage{times}  
\usepackage{helvet}  
\usepackage{courier}  
\usepackage[hyphens]{url}  
\usepackage{graphicx} 
\usepackage{natbib}  
\usepackage{caption} 
\usepackage{algorithm}
\usepackage{algorithmic}

\usepackage{newfloat}
\usepackage{listings}
\DeclareCaptionStyle{ruled}{labelfont=normalfont,labelsep=colon,strut=off} 
\floatstyle{ruled}
\newfloat{listing}{tb}{lst}{}
\floatname{listing}{Listing}
\title{SMamba: Sparse Mamba for Event-based Object Detection}
\author{
    Nan~Yang\textsuperscript{\rm 1}\equalcontrib,
    Yang~Wang\textsuperscript{\rm 1}\equalcontrib,
    Zhanwen~Liu\textsuperscript{\rm 1}\thanks{Corresponding author.},
    Meng~Li\textsuperscript{\rm 2},
    Yisheng~An\textsuperscript{\rm 1},
    Xiangmo~Zhao\textsuperscript{\rm 1}
}
\affiliations{
    \textsuperscript{\rm 1}School of Information Engineering, Chang'an University, China\\
    \textsuperscript{\rm 2}School of Civil Engineering, Tsinghua University, China\\

    \{2022024001, ywang120, zwliu, aysm, xmzhao\}@chd.edu.cn, mengli@tsinghua.edu.cn
}

\usepackage{bibentry}

\begin{document}

\maketitle

\begin{abstract}
Transformer-based methods have achieved remarkable performance in event-based object detection, owing to the global modeling ability. However, they neglect the influence of non-event and noisy regions and process them uniformly, leading to high computational overhead. To mitigate computation cost, some researchers propose window attention based sparsification strategies to discard unimportant regions, which sacrifices the global modeling ability and results in suboptimal performance. To achieve better trade-off between accuracy and efficiency, we propose Sparse Mamba (SMamba), which performs adaptive sparsification to reduce computational effort while maintaining global modeling capability. Specifically, a Spatio-Temporal Continuity Assessment module is proposed to measure the information content of tokens and discard uninformative ones by leveraging the spatiotemporal distribution differences between activity and noise events. Based on the assessment results, an Information-Prioritized Local Scan strategy is designed to shorten the scan distance between high-information tokens, facilitating interactions among them in the spatial dimension. Furthermore, to extend the global interaction from 2D space to 3D representations, a Global Channel Interaction module is proposed to aggregate channel information from a global spatial perspective. Results on three datasets (Gen1, 1Mpx, and eTram) demonstrate that our model outperforms other methods in both performance and efficiency.
\begin{links}
\link{Code}{https://github.com/Zizzzzzzz/SMamba_AAAI2025}
\end{links}
\end{abstract}

%

\section{Introduction}

\begin{figure}[t]
\centering
\includegraphics[scale=1.4]{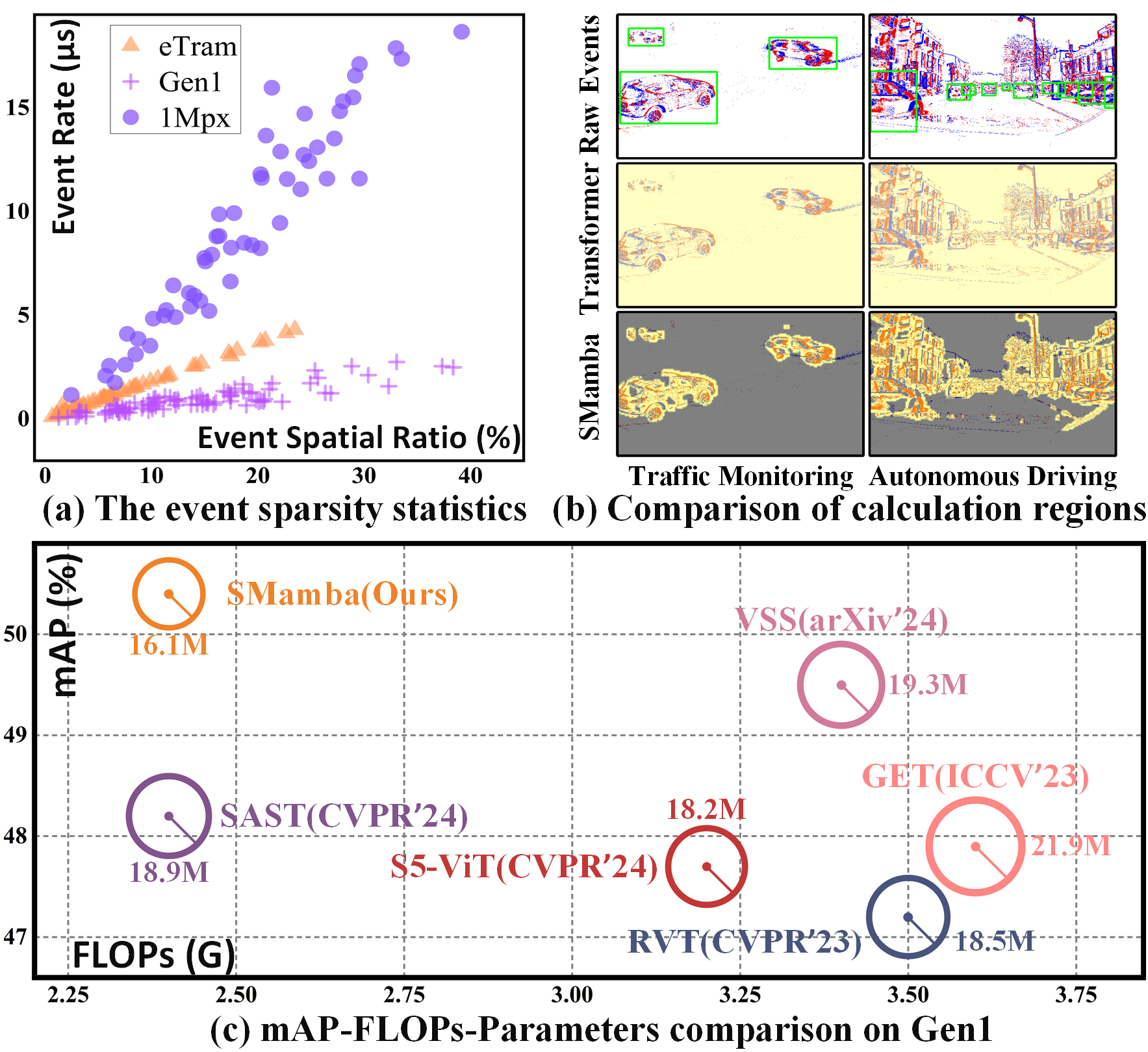}
\caption{(a) The event sparsity statistics on three datasets. Each point represents a scenario. (b) Compared with calculating all regions uniformly in Transformer, our proposed SMamba can suppress non-event and noisy regions (gray regions) from participating in calculations and only retain information-rich areas (yellow regions), alleviating computational overhead significantly and suppressing noise disturbance, simultaneously. (c) mAP-FLOPs-Parameters comparison between state-of-the-art methods and our SMamba on Gen1, where the circle radius is the parameter. SMamba achieves a superior balance between accuracy and efficiency.}
\label{fig111}
\end{figure}

Robust object detection is fundamental for intelligent systems, such as autonomous driving and robotics \cite{liu2024multi,zhang2013multi,li2024intention,liu2023multi}. However, frame-based cameras encounter inherent limitations in frame rate and dynamic range, leading to low-quality images and hindering the extraction of discriminative features in challenging scenarios, such as high-speed motion and adverse exposure conditions (\emph{e.g.}, low light and over-exposure) \cite{4444573,sayed2021improved, liu2024boosting,yan2018two,liu2024enhancing}. In these scenarios, event cameras emerge as a superior alternative due to their unique advantages, which asynchronously detect pixel-level changes in light intensity, offering high temporal resolution and high dynamic range, exhibiting stable and robust performance under challenging conditions \cite{finateu20205, huang2023event, son20174, gallego2020event}.

To fully exploit the superior performance of event cameras in object detection task, researchers have developed various methods based on well-designed neural network architectures, including SNN-based, GNN-based, CNN-based, and Transformer-based methods. Theoretically, SNN-based \cite{cordone2022object} and GNN-based \cite{schaefer2022aegnn} methods can achieve low-latency inference, but require specialized hardware and perform poorly in real scenarios. Recent research indicates that Transformer-based \cite{gehrig2023recurrent,peng2023get,zubic2024state, Zubic_2023_ICCV} methods, benefiting from their larger receptive field, significantly outperform local receptive field-based CNN methods \cite{dosovitskiy2020vit,NEURIPS2020_c2138774, li2022asynchronous}. However, these methods typically transform event streams into discrete tokens and process non-event and noisy regions uniformly, neglecting the influence of spatial sparsity \cite{peng2024scene} and low signal-to-noise ratio \cite{ding2023mlb}, leading to substantial redundant computation and suboptimal performance. As shown in Figure \ref{fig111}(a), we calculate the ratio of triggered event pixels (event spatial ratio) and the average number of events over the camera's operating time (event rate) for the Gen1 \cite{de2020large}, 1Mpx \cite{NEURIPS2020_c2138774}, and eTram \cite{verma2024etram} datasets. The results indicate that in 96\% of the scenes, the event spatial ratio is below 30\%, indicating high spatial sparsity. To improve the computation efficiency on spare event data, SAST \cite{peng2024scene} proposes the window attention based sparsification strategy to discard unimportant regions but sacrifices the global modeling capability. Additionally, the token scoring module is optimized through an indirect gradient propagation path, complicating the optimization of the framework and necessitating more iterations.

In this paper, we propose Sparse Mamba (SMamba) which effectively reduces the computational cost by adaptively discarding uninformative tokens, and captures global context through information-guided spatial selective scanning and global spatial-based channel selective scanning, achieving better accuracy and efficiency trade-off, as depicted in Figure \ref{fig111}(b, c). Specifically, we design the Spatio-Temporal Continuity Assessment (STCA) module by leveraging the spatiotemporal distribution differences between activity and noisy events, without complex module designs and challenging optimizations. The information content of tokens is measured by evaluating the spatiotemporal continuity of events within each token's corresponding region, generating a sparsification map to guide the non-event and noisy tokens discarding. Based on the assessment results, we propose the Information-Prioritized Local Scanning (IPL-Scan) strategy, which sorts tokens at the window level based on their information content, facilitating interactions among high-information tokens in 2D space while preserving critical local content. Additionally, to further extend the global interaction from 2D space to 3D representations, we propose the Global Channel Interaction (GCI) module to perform dynamic integration of channel information based on 2D global context. The experimental results on three datasets (Gen1, 1Mpx, and eTram) demonstrate that our method achieves superior performance and greater efficiency. Overall, our contribution can be summarized as follows:

(1) We propose the SMamba, which adaptively discards non-event and noisy tokens based on the spatiotemporal continuity assessment and captures the global relationship across spatial and channel dimensions, demonstrating an optimal balance between accuracy and efficiency.

(2) The IPL-Scan is designed to guide the model to focus on high-information tokens during the scanning process, thereby improving spatial contextual modeling capability.

(3) The GCI module is devised to extend global interactions into the 3D feature space by aggregating channel information from the global spatial perspective, further improving global modeling capability.

(4) Experimental results on Gen1, 1Mpx, and eTram datasets demonstrate that our SMamba surpasses state-of-the-art methods, achieving superior performance.

\section{Related Work}

This section provides a comprehensive overview of event-based object detection methods, followed by a review of recent advancements in Vision Mamba.

\subsection{Event-Based Object Detection}

Existing event-based object detection methods can be categorized according to the neural network architecture employed: SNN-based, GNN-based, CNN-based, and Transformer-based methods. SNN-based \cite{cordone2022object} and GNN-based \cite{schaefer2022aegnn} methods effectively leverage the sparse and asynchronous properties of event data, thereby achieving high computational efficiency. However, the immaturity of their network architectures constrains their capacity to handle complex object detection task, leading to limited performance.

CNN-based and Transformer-based methods initially convert events into image-like tensors, facilitating compatibility with subsequent processing architectures. Prominent CNN-based methods, such as RED \cite{NEURIPS2020_c2138774} and ASTMNet \cite{li2022asynchronous}, integrate CNNs with RNNs to effectively extract spatiotemporal features. Recently, Transformer-based methods \cite{gehrig2023recurrent, peng2023get, zubic2024state} benefit from the global receptive field of self-attention mechanism, surpassing CNN-based methods and achieving outstanding performance. However, the self-attention mechanism uniformly processes non-event and noisy regions, resulting in considerable redundant computation. To address this, SAST \cite{peng2024scene} proposes a window-token co-sparsification strategy that adaptively discards insignificant windows and tokens. Nevertheless, this method relies on window attention to achieve low computational overhead, sacrificing global modeling capability.

\begin{figure*}[t]
\centering

\includegraphics[scale=0.719]{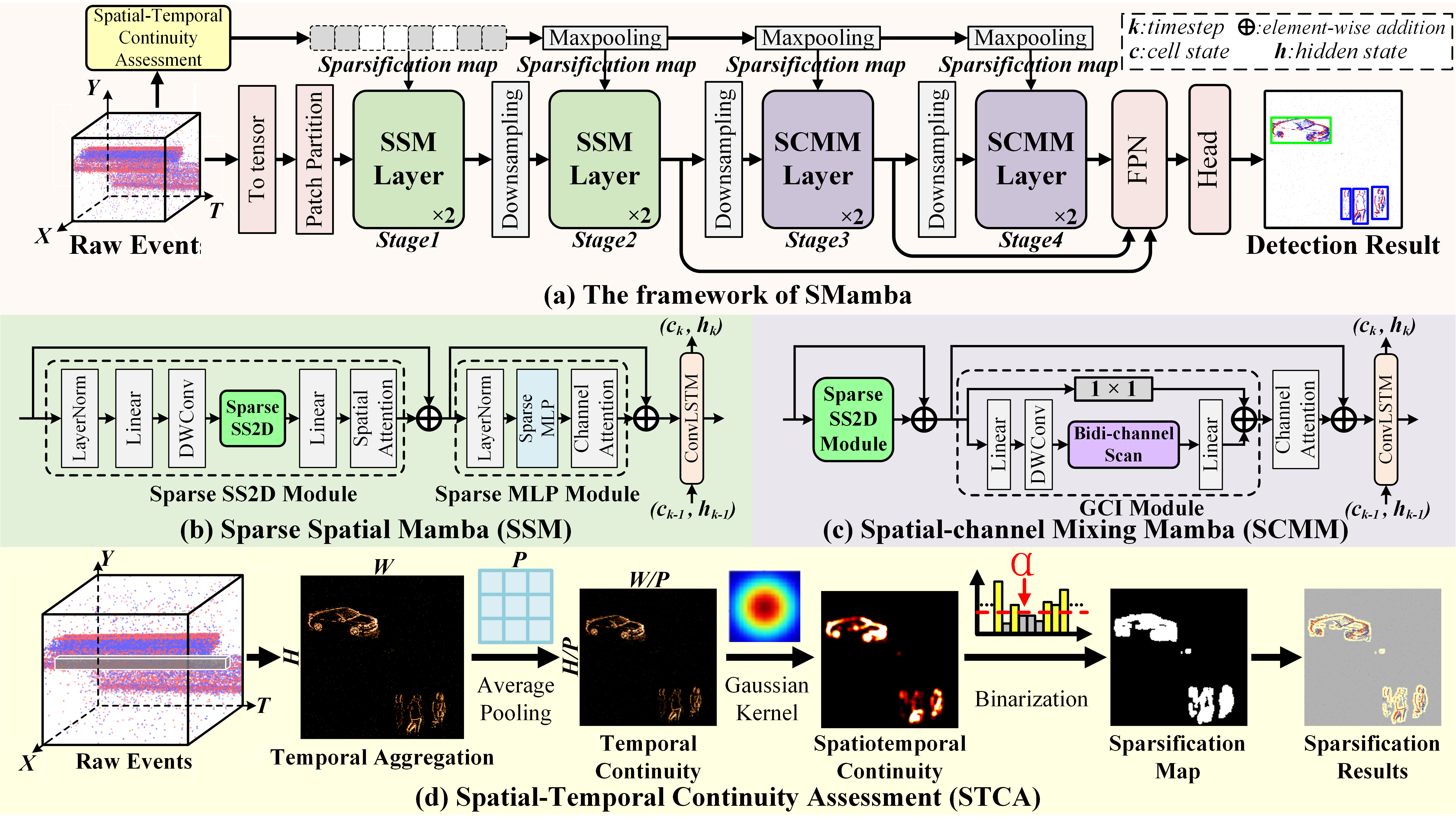}
\caption{\textbf{The architecture of SMamba}. Given the input event stream, the STCA module (as illustrated in (d)) evaluates the information content of tokens based on spatiotemporal continuity and generates a sparsification map to guide the sparsification operation. Simultaneously, the event stream is preprocessed and transferred to four stages for comprehensive global interactions across spatial and channel dimensions. The first two stages employ the Sparse Spatial Mamba (SSM) layer (as detailed in (b)) to facilitate global spatial interactions on kept tokens. The final two stages utilize the Spatial-Channel Mixing Mamba (SCMM) layer (as depicted in (c)) to permute global modeling within the 3D representation space.}
\label{fig2}
\end{figure*}

\subsection{Vision Mamba}

Mamba achieves a better balance between efficiency and performance, making it an effective alternative to Transformer. Specifically, Mamba improves global modeling capability by introducing an input-dependent selective scanning mechanism (S6) and proposes a parallel scanning mechanism to retain the linear complexity of State Space Models (SSMs). Inspired by Mamba's success in the NLP field, Vim \cite{zhu2024vision} and Vamba \cite{liu2024vmamba} integrate Mamba into the design of vision backbone networks, proposing multi-scan strategies to adapt to the non-causal nature of images, achieving groundbreaking results. Subsequently, numerous studies have applied Mamba to various vision tasks, including medical image segmentation \cite{xu2024survey}, remote sensing image segmentation \cite{patro2024mamba}, image restoration \cite{guo2024mambair,zou2024wave}, and object detection and tracking \cite{huang2024mamba, dong2024fusion}, demonstrating Mamba's significant potential in the visual domain. However, Mamba lacks effective mechanisms for handling sparse event data, resulting in substantial redundant computations.

\section{Method}
\begin{figure*}[tbp]
\centering

\includegraphics[scale=2.5]{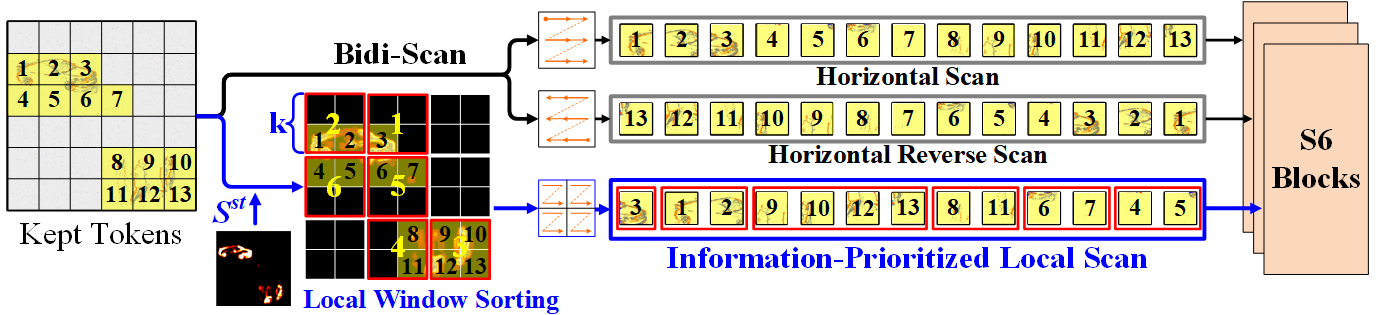}
\caption{\textbf{Sparse SS2D}. For ease of observation, a large patch size is utilized for tokenization. The uninformative and noisy tokens (gray regions) are discarded from the calculation and the kept tokens (yellow regions) are expanded into three scanning sequences by Bidi-Scan and IPL-Scan, each sequence is processed in parallel using separate S6 blocks. The window sorting results of IPL-Scan are indicated at the center of each window (highlighted in red).}
\label{fig3}
\end{figure*}

\subsection{Preliminary: SSMs and Mamba}

The classical State Space Model (SSM) is a linear time-invariant system. Given an input sequence $x(t)\in{\mathbb{R}^L}$, it maintains a hidden state $h(t)\in{\mathbb{R}^N}$ to store the contextual information and generates an output $y(t)\in{\mathbb{R}^L}$. The computational procedure is as follows:
\begin{equation}
h^{\prime}(t)  =\mathbf{A} h(t)+\mathbf{B} x(t), y(t)  =\mathbf{C} h(t),
\end{equation}
where $A\in{\mathbb{R}^{N \times N}}$ is the state matrix, $B\in{\mathbb{R}^{N \times L}}$ and $C\in{\mathbb{R}^{L \times N}}$ are the input and output matrix, respectively.

To improve the global modeling ability of SSM, Mamba \cite{mamba} proposes the input-dependent selective scanning mechanism (S6) which introduces nonlinearity and discretization by incorporating a time scale parameter $\Delta$ and the zero-order hold (ZOH). The procedure is as follows:
\begin{equation}
h_{t}  =\overline{\mathbf{A}} h_{t-1}+\overline{\mathbf{B}} x_{t}, y_{t}  =\mathbf{C} h_{t},
\end{equation}
\begin{equation}
\overline{\mathbf{A}}=\exp(\Delta\mathbf{A}),\overline{\mathbf{B}}=(\Delta\mathbf{A})^{-1}(\exp(\Delta\mathbf{A})-\mathbf{I})\cdot\Delta\mathbf{B}.
\end{equation}

\subsection{The Overview of the SMamba}

The framework of SMamba is illustrated in Figure \ref{fig2}(a). Initially, the event stream is input into the Spatial-Temporal Continuity Assessment (STCA) module, depicted in Figure \ref{fig2}(d), which generates a sparsification map to guide the sparsification operations. Meanwhile, the event stream is converted into voxel tensors \cite{zhu2019unsupervised} and divided into patches for tokenization. These tokens are then processed through four stages for multi-scale feature extraction. The first two stages employ the Sparse Spatial Mamba (SSM) layer, as shown in Figure \ref{fig2}(b), which includes a Sparse SS2D (2D Selective Scan) module to improve global spatial interactions on kept tokens, a Sparse MLP module to further reduce computational overhead, and a ConvLSTM \cite{shi2015convolutional} to transmit spatiotemporal information across time steps, with its output sent to subsequent layers. The final two stages utilize the Spatial-Channel Mixing Mamba (SCMM) layer, as shown in Figure \ref{fig2}(c), which comprises a Sparse SS2D module, a Global Channel Interaction (GCI) module which extends the global modeling to 3D representations by facilitating channel interactions from a global perspective, and a ConvLSTM. Features produced by the last three stages are then fed into a Feature Pyramid Network (FPN) for multi-scale feature fusion. Finally, the YOLOX \cite{yolox2021} detection head outputs the detection results.

\subsection{Spatial-Temporal Continuity Assessment}
Event cameras asynchronously trigger events at locations where brightness changes exceed the threshold, resulting in significant spatial sparsity, particularly in scenarios where the camera is stationary \cite{verma2024etram}. Furthermore, the inherent circuitry characteristics of event cameras generate substantial noise \cite{ding2023mlb, duan2024led}. These blank and noisy regions are uninformative, leading to unnecessary computations and potential interference.

We observe that activity events and noise events exhibit significant differences in their spatiotemporal distribution. Specifically, the noise events are spatially isolated or temporally discontinuous, whereas activity events are typically localized at the edges of moving objects, exhibiting spatial proximity and temporal continuity \cite{kim2021n}. Based on this prior, we propose the Spatial-Temporal Continuity Assessment (STCA) module, as depicted in Figure \ref{fig2}(d), which evaluates token importance and selectively discards uninformative tokens by assessing spatiotemporal continuity of events, reducing computational overhead.

Specifically, given the event stream $\left\{\left(x_{i}, y_{i}, t_{i}, p_{i}\right)\right\}_{i=1}^{N}$, where $(x_i,y_i)$ is the spatial coordinates, $t_i$ represents the timestamp, and $p_i\in\{{-1,1}\}$ indicates the event polarity. The timestamps of events at each pixel are first accumulated to generate the temporal continuity score map $S^{t}\in{\mathbb{R}^{H \times W}}$, which quantifies the temporal continuity at each spatial location. The formulation is as follows:

\begin{equation} 
{S^{t}_{x,y}} = {\sum\limits_{{i,x_{i}=x,y_{i}=y}} t_{i}}.
\end{equation}

Next, average pooling with kernel size and stride of $P$ is employed to extract the temporal information content $S^{t}\in{\mathbb{R}^{H/P \times W/P}}$ corresponding to each token, where $P$ represents the patch size utilized during event tokenization. Subsequently, the neighborhood information is effectively aggregated to assess spatial continuity. For activity events, nearby neighbors are more likely to be triggered by the same moving edge, while distant neighbors are more likely to be noise. Therefore, to mitigate the impact of noise on information content evaluation, a Gaussian function is employed to perform distance-based weighted aggregation within the neighborhood, thereby smoothing the noise while maintaining more complete object structures \cite{wan2022s2n}. The formulation is as follows:

\begin{equation} 
{S^{st}} = \frac{\sum_{q \in \Omega}\left( \exp \left(-\frac{\|q-c\|^{2}}{2 \sigma^{2}}\right) S^{t}_{q}\right)}{\sum_{q \in \Omega} \exp \left(-\frac{\|q-c\|^{2}}{2 \sigma^{2}}\right)},
\end{equation}
where $c$ is the center of neighborhood $\Omega$, $S^{t}_{q}$ denotes the value of neighbor $q$, and $\sigma$ represents the variance. In the resulting spatiotemporal continuity score map $S^{st}$, each pixel value indicates the activity event information content of the token. The larger the value, the more important the token.

The mean value of the spatiotemporal continuity score map $S^{st}$ indicates the sparsity of the scene and serves as a threshold for discarding uninformative tokens. To adaptively retain important tokens based on the scene's sparsity level and avoid the loss of critical object information, a manually adjusted sparsity scaling factor $\beta$ is introduced to modulate the discard ratio. The threshold $\alpha$ is defined as follows:

\begin{equation}
{\alpha} = \frac {sum \left(S^{st}\right)} {\beta \frac{HW}{P^{2}}}.
\end{equation}

Based on this threshold, a sparsification map $D\in{\mathbb{R}^{H/P \times W/P}}$ is generated for sparsification. The expression is as follows:
\begin{equation} 
{D_{x,y}} = \left\{ {\begin{array}{*{20}{l}}
{ 1, \; {\rm{ if }} \;  {S^{st}_{x,y}}  >= \alpha,}\\
{ 0, \; {\rm{ if }} \; {S^{st}_{x,y}}  < \alpha.}
\end{array}} \right.
\end{equation}

The sparsification map is subsequently propagated to the following layers to guide sparsification operations.

\subsection{Information-Prioritized Local Scan}
The 2D spatial scanning strategies, such as Bidi-Scan and Cross-Scan, may disperse tokens associated with the same object in the scanning sequence, resulting in distant scanning intervals and weakened interactions between them \cite{liu2024vmamba, shi2024multi}. Therefore, we propose the Information-Prioritized Local Scan (IPL-Scan) which mitigates the limitations of 2D scan methods, and design the Sparse SS2D which combines IPL-Scan and Bidi-Scan to facilitate global interactions, as depicted in Figure \ref{fig3}.

The spatiotemporal continuity score map quantifies the information of tokens, with higher scores indicating a higher likelihood of being a foreground object. Reordering tokens according to this map, tokens with higher informational content are processed earlier, which shortens the scan distance between important tokens, facilitating interactions among them. Furthermore, tokens with lower scores are processed later, which mitigates potential interference from noise.

Considering that the direct reordering might disrupt local information, a local constraint is introduced to the sorting process. When a token is processed, its $k \times k$ neighborhood is also processed immediately afterward. Specifically, max pooling with a kernel and stride of $k$ is used to extract the maximum value from each $k \times k$ local window. These maximum values, representing the local windows, are initially sorted. Subsequently, the sorted results are upsampled by $k$ to obtain the window-level sorted outcome. This strategy effectively facilitates the interaction among potential object regions while preserving local information.

\subsection{Global Channel Interaction}

To extend the global interaction from 2D space to 3D representations, we propose the Global Channel Interaction (GCI) module, as shown in Figure \ref{fig2}(c), which integrates the Bidirectional Channel (Bidi-channel) Scan mechanism with $1\times1$ convolution to perform dynamic integration of channel information based on global and local content.

\begin{figure}[tbp]
\centering
\includegraphics[scale=0.8]{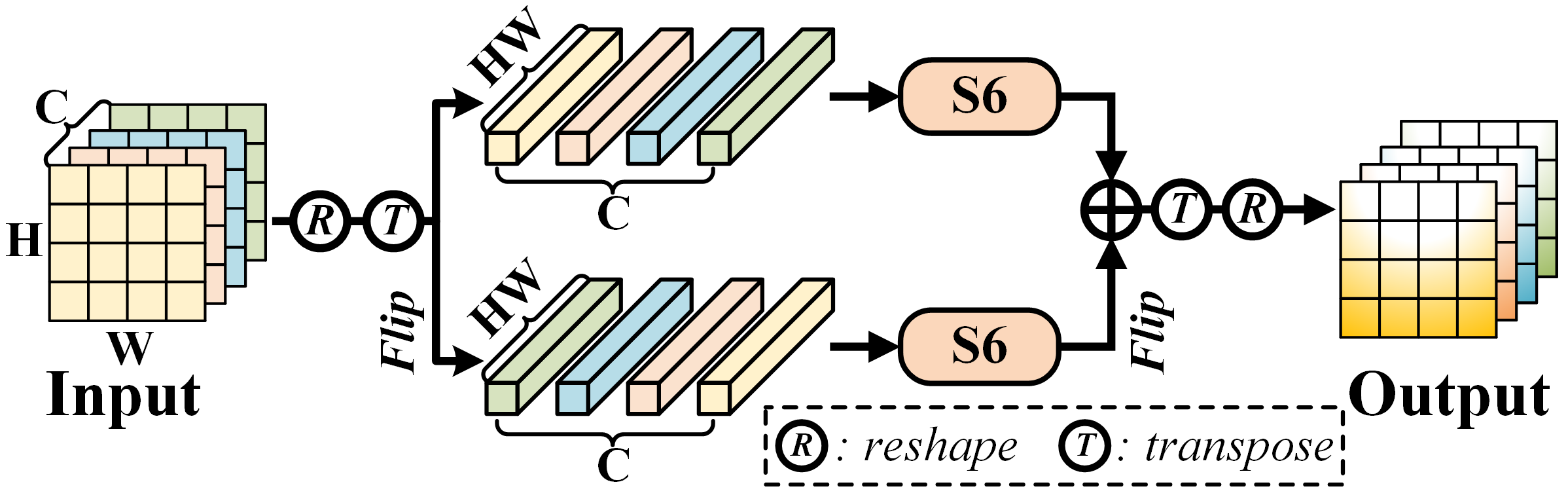}
\caption{\textbf{Bidirectional Channel Scan.} The feature map is unfolded along the spatial dimension and arranged into two sequences along the channel direction, which are then input into S6 for full spatial-level channel information interaction.}
\label{fig4}
\end{figure}

The feature tensor $X\in{\mathbb{R}^{C \times H \times W}}$ is processed through two distinct branches: the Bidi-channel Scan and the $1\times1$ convolution, facilitating channel interactions at global and local levels. Within the Bidi-channel Scan branch, $X$ is preprocessed by Linear and DWConv (Depthwise Convolution) to capture the local context and then sent into the Bidi-channel Scan, as shown in Figure \ref{fig4}. $X$ is flattened along the $H$ and $W$ dimensions, followed by a transposition to treat the global spatial information of each channel as the fundamental unit of interaction. Subsequently, a reverse sequence is generated by flipping, which is then input into S6 along with the original sequence to enable adaptive interaction from the global perspective. Performing selective scanning based on global spatial content allows each channel to selectively focus on other channels from a more comprehensive perspective, accurately capturing dependencies between channels and further enhancing global modeling capability. The other branch employs $1\times1$ convolution to capture pixel-level dependencies between channels, enabling local adaptive interaction. Finally, the results from both branches are integrated to achieve comprehensive channel interactions.

\begin{table*}[htbp]
\small
\centering

\begin{tabular}{c|c|cccc|cccc}
     
    \Xhline{1.5pt}
    {\multirow{2}*{{\textbf{Methods}}}} & {\multirow{2}*{{\textbf{Backbone}}}} & \multicolumn{4}{c}{{\textbf{Gen1}}} & \multicolumn{4}{c}{{\textbf{1Mpx}}}\\
    \hhline{~~--------}
     & & mAP & FLOPs & Params & Runtime & mAP & FLOPs & Params & Runtime\\
    \hline
    \hline
    RED & CNN+RNN & 40.0 & 6.0G & 24.1M & 16.7ms & 43.0 & 19.0G &24.1M & 39.3ms \\
    ASTMNet & CNN+RNN & 46.7 & 29.3G & \textgreater 100M & 35.6ms & 48.3 & 75.7G & \textgreater 100M & 72.3ms \\
    \hline
    \hline
    ERGO-12 & Transformer & \textbf{50.4} & 50.8G & 59.6M & 69.9ms & 40.6 & 50.8G & 59.6M &100ms \\
    {RVT-B} & { Transformer+RNN} & 47.2 & 3.5G & 18.5M & 10.2ms & 47.4 & 10.3G & 18.5M & 11.9ms \\
    GET-T & { Transformer+RNN} & 47.9 & 3.6G & 21.9M & 16.8ms & 48.4 & 10.6G & 21.9M & 21.9ms \\
    SAST-CB & { Transformer+RNN} & 48.2 & 2.4G & 18.9M & 22.7ms& 48.7 & 6.4G & 18.9M &23.6ms \\
    S5-ViT-B & { Transformer+SSM} & 47.7 & \textgreater 3.1G & 18.2M & 9.4ms & 47.8 & \textgreater 9.1G & 18.2M & 10.9ms \\
    \hline
    \hline
    VSS & SSM+RNN & 49.5 & 3.4G & 19.3M & 17.4ms & 48.2 & 10.3G  & 19.3M & 17.7ms \\
    Baseline & SSM+RNN & \underline{50.0} & 3.1G & 16.1M & 25.2ms & \underline{48.8} & 9.5G & 16.7M & 27.5ms \\
    SMamba & SSM+RNN & \textbf{50.4} & 2.4G (-23\%) & 16.1M & 24.0ms& \textbf{49.3} & 7.4G (-22\%) & 16.7M & 26.0ms \\
    \Xhline{1.5pt}
\end{tabular}
\caption{Performance compared with SOTA methods on two autonomous driving datasets Gen1 and 1Mpx. The reported FLOPs belongs to the backbone. Values in brackets (·) indicate the percentage decrease in FLOPs compared to the baseline method.}
\label{table1}
\end{table*}

\section{Experiments}
This section begins with an overview of the experimental setup. Subsequently, a comparative analysis of our method against state-of-the-art (SOTA) methods is presented. The visualization results are then shown to demonstrate the scene-adaptability of our method. Finally, ablation studies are conducted to validate the effectiveness of our approach.

\subsection{Experimental Setup}

This subsection details the employed datasets, the validation metrics utilized, and the implementation details.

\textbf{Datasets.} We conduct experiments on two autonomous driving datasets Gen1 \cite{de2020large} and 1Mpx \cite{NEURIPS2020_c2138774}, and one traffic monitoring dataset eTram \cite{verma2024etram}. The Gen1 dataset comprises over 39 hours of event data at 304 × 240 resolution, providing more than 255,000 labeled cars and pedestrians with annotation frequencies of 1 Hz, 2 Hz, or 4 Hz. The 1Mpx dataset provides 14.65 hours of data at a higher resolution of 1280 × 720 and 60 Hz annotated frequency, containing over 25 million labeled boxes across seven categories. The eTram dataset, collected for traffic monitoring, includes approximately 10 hours of data at 1280 × 720 resolution, encompassing around 2 million labeled boxes across eight categories with an annotation frequency of 30 Hz. The eTram is collected from a roadside perspective and exhibits greater sparsity compared to the other two datasets due to the camera's fixed position \cite{verma2024etram}.

\textbf{Metrics.} The COCO mAP (mean average precision) \cite{lin2014microsoft} is used to evaluate the accuracy of object detection. The model's size is measured by the parameter count. Additionally, following SAST \cite{peng2024scene}, we compute the average FLOPs (Floating Point Operations Per Second) on the first 1,000 samples of the test set to assess the computational complexity. We also compare the inference time (runtime) with other methods.

\textbf{Implementation Details.} To guarantee comparison fairness, we follow the dataset preprocessing methods, augmentation techniques, mixed batching strategy, event representation method and evaluation protocols established in RVT \cite{gehrig2023recurrent}.

\subsection{Quantitative Results}
We provide a comparative analysis of our method against 2 CNN-based methods: RED \cite{NEURIPS2020_c2138774}, ASTMNet \cite{li2022asynchronous}; as well as 5 Transformer-based methods: ERGO-12 \cite{Zubic_2023_ICCV}, RVT \cite{gehrig2023recurrent}, GET \cite{peng2023get}, SAST \cite{peng2024scene} and S5-ViT \cite{zubic2024state} on the Gen1, 1Mpx datasets. On the eTram dataset, we compare our method with 3 Transformer-based methods: RVT, SAST and S5-ViT, as the relevant code from other works has not been released. To compare with SSM-based methods, we employ the VSS block from VMamba \cite{liu2024vmamba} to construct a detection framework named VSS. Moreover, a baseline model without the sparsification strategy is established to evaluate the effectiveness of the proposed method. 

The results are shown in Table \ref{table1} and Table \ref{table11}. On the Gen1 dataset, our SMamba outperforms all other methods with the lowest FLOPs and parameter count. Compared to ERGO-12, SMamba achieves the same mAP with merely 5\% of its FLOPs and 27\% of its parameter count. On the 1Mpx and eTram datasets, SMamba outperforms SAST-CB by 0.6\% and 2.6\% in mAP with similar FLOPs and lower parameter counts. By further integrating our sparsification strategy into our baseline, SMamba reduces FLOPs by 23\%, 22\% and 31\%, along with mAP gains of 0.4\%, 0.5\% and 0.3\% in three datasets, respectively. Our sparsification operation imposes network focus on important areas, alleviating interference from blank and noisy regions, thereby reducing computational overhead and improving accuracy. The inference speed of SMamba is faster than CNN-based methods and Transformer-based method ERGO-12, and is comparable with SAST-CB while achieving higher accuracy. The consistent performance improvements on the autonomous driving and traffic surveillance datasets demonstrate that our method can generalize to varying sparsity levels while achieving an ideal trade-off between accuracy and efficiency.

     

\begin{table}[t]
\small
\centering
\begin{tabular}{c|cccc}
     
    \Xhline{1.5pt}
    {\multirow{2}*{{\textbf{Methods}}}}  & \multicolumn{4}{c}{{\textbf{eTram}}}\\
    \hhline{~----}
      & mAP & FLOPs & Params & Runtime\\
    \hline
    \hline
    RVT-B   & 29.5 & 10.3G & 18.5M & 11.9ms \\
    SAST-CB   & 30.0 & 6.2G & 18.9M & 24.4ms \\
    S5-ViT-B  & 29.3 & \textgreater 9.1G & 18.2M & 10.9ms \\
    \hline
    \hline
    VSS  & 31.3 & 10.3G & 19.3M & 17.7ms \\
    Baseline   & \underline{32.3} & 9.5G & 16.7M & 27.5ms \\
    SMamba  & \textbf{32.6} & 6.6G (-31\%) & 16.7M & 25.2ms \\
    \Xhline{1.5pt}
\end{tabular}
\caption{Performance compared with SOTA methods on the traffic monitoring dataset eTram.}
\label{table11}
\end{table}

\begin{table}[t]
\small
\centering
\begin{tabular}{c|c|c|c}
     
    \Xhline{1.5pt}
    \textbf{Methods} & mAP & FLOPs & Params\\
    \hline
    Variance & 30.8 & 6.5G & 16.7M\\
    Entropy & 30.4 & 6.6G & 16.7M\\
    Scoring module & 31.2 & 7.0G & 17.0M\\
    STCA & \textbf{32.6} & 6.6G & 16.7M\\
    \Xhline{1.5pt}
\end{tabular}
\caption{Performance of different scoring methods.}
\label{table2}
\end{table}

\begin{figure}[t]
\centering

\includegraphics[scale=2.0]{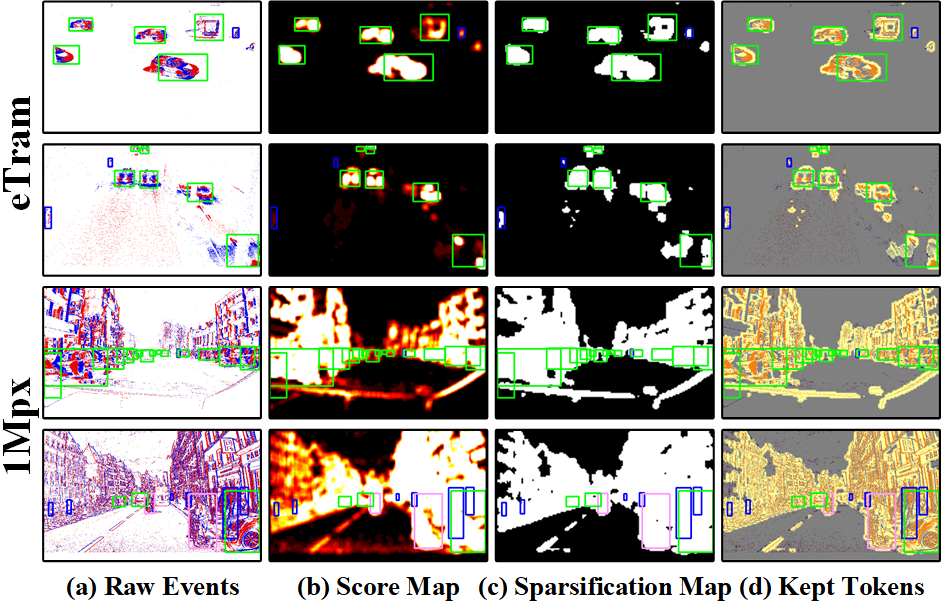}
\caption{Visualizations of raw events, score map, sparsification map and sparsification results.}
\label{fig5}
\end{figure}

\subsection{Sparsification Visualizations}

The visualizations of raw events, score map, sparsification map, and sparsification results on the eTram and 1Mpx datasets are shown in Figure \ref{fig5}, with increasing levels of scene complexity. The eTram, collected by a stationary camera, exhibits greater sparsity compared to the 1Mpx which is obtained from a moving camera. As event density increases, the STCA module keeps an increasing number of tokens. This indicates that our STCA exhibits robust scene-adaptive capability, effectively mitigating interference from blank areas and noise while selecting important tokens.

\subsection{Ablation Studies}

To assess the effectiveness of the proposed method, we conduct a series of ablation studies on the eTram dataset.

\textbf{STCA Module.} We compare our STCA with two information content evaluation metrics, variance and entropy, as well as a learnable event scoring module \cite{peng2024scene} while maintaining architectural consistency. As shown in Table \ref{table2}, our method exhibits superior performance, surpassing other methods. Variance and entropy cannot distinguish between activity events and noise events, leading to misjudgments of information content. The scoring module is optimized through an indirect gradient propagation path, resulting in suboptimal token scoring. In contrast, the STCA effectively distinguishes activity events from noise and blank areas based on the spatiotemporal continuity prior.

\begin{table}[tb]
\small
\centering
\begin{tabular}{c|c|c|c}
     
    \Xhline{1.5pt}
    \textbf{Methods} & mAP & FLOPs & Params\\
    \hline
    IPL-Scan & 29.9 & 6.5G & 16.3M\\
    Bidi-Scan & 30.2 & 6.6G & 16.5M\\
    Cross-Scan & 30.7 & 6.6G & 16.9M\\
    Bidi-Scan + IPL-Scan & \textbf{32.6} & 6.6G & 16.7M\\
    \thead{Bidi-Scan + IPL-Scan \\ (w/o local constraint)} & 30.7 & 6.6G & 16.7M\\
    
    \Xhline{1.5pt}
\end{tabular}
\caption{Performance of different scanning patterns.}
\label{table3}
\end{table}

\begin{table}[t]
\small
\centering
\begin{tabular}{cc|c|c|c}
     
    \Xhline{1.5pt}
    \textbf{Bidi-channel Scan} & \textbf{$1 \times 1$} & mAP & FLOPs & Params\\
    \hline
    \multicolumn{2}{c|}{MLP} & 31.0 & 9.7G & 19.2M\\
    \hdashline
    {\checkmark} &   & 31.5 & 6.3G & 16.0M\\
    {\checkmark} & {\checkmark}  & \textbf{32.6} & 6.6G & 16.7M\\
    
    \Xhline{1.5pt}
\end{tabular}
\caption{Performance of the components of the GCI module.}
\label{table4}
\end{table}

\textbf{Scanning Pattern.} We perform an ablation on the scanning pattern within the Sparse SS2D, the results are shown in Table \ref{table3}. Increasing the 2D scanning paths from 2 (Bidi-Scan) to 4 (Cross-Scan) improves the mAP by merely 0.5\%. Employing IPL-Scan alone yields the poorest performance due to the significant loss of spatial structure information. However, the combination of Bidi-Scan and IPL-Scan achieves the best performance, increasing the mAP by 2.4\%. Removing the local constraint in the IPL-Scan results in a significant decline in performance, indicating the crucial role of local information.

\begin{figure}[t]
\centering
\includegraphics[scale=1.98]{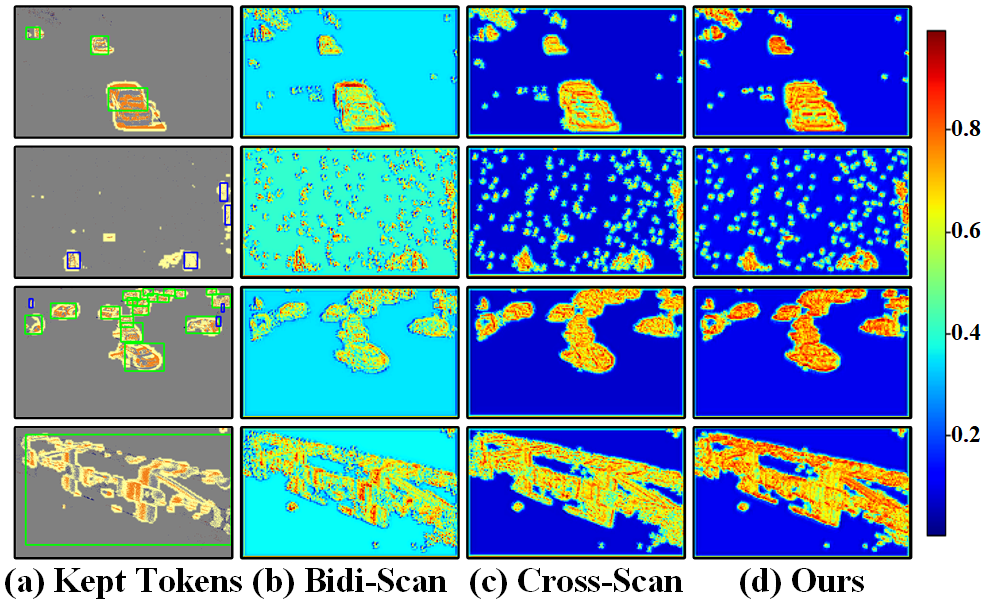}
\caption{The feature visualization of Bidi-Scan, Cross-Scan and our method.}
\label{fig6}
\end{figure}

Figure \ref{fig6} presents a visual comparison of our scan method with Bidi-Scan and Cross-Scan. Bidi-Scan and Cross-Scan organize scanning sequences in the 2D direction, which limits contextual interaction, resulting in less distinct features. In contrast, our IPL-Scan shortens the scanning distance between different regions of the same object, thereby facilitating interaction among these regions and enabling the modeling of more discriminative feature representations.



\begin{table}[t]
\small
\centering
\begin{tabular}{cccc|c|c|c}
     
    \Xhline{1.5pt}
    \textbf{S1} & \textbf{S2} & \textbf{S3} & \textbf{S4} & mAP & FLOPs & Params\\
    \hline
     \multicolumn{4}{c|}{MLP} &  31.0 & 9.7G & 19.2M\\
     \hdashline
     &  &  & {\checkmark} &  31.0 & 6.2G & 16.7M\\
     &  & {\checkmark} & {\checkmark} &  \textbf{32.6} & 6.6G & 16.7M\\
     & {\checkmark} & {\checkmark} & {\checkmark} &  31.4 & 6.8G & 17.2M\\
    {\checkmark} & {\checkmark} & {\checkmark} & {\checkmark} &  31.5 & 7.0G & 17.8M\\
    
    \Xhline{1.5pt}
\end{tabular}
\caption{Performance of GCI placed after different stages.}
\label{table5}
\end{table}

\textbf{Global Channel Interaction Module Design.} We replace the GCI module with MLP as the baseline to analyze the contribution of Bidi-channel Scan and $1\times1$ convolution. The results are shown in Table \ref{table4}, replacing MLP with Bidi-channel Scan improves mAP by 0.5\% while reducing FLOPs and parameters by 35\% and 16.7\%,  respectively. The global spatial context provides a more comprehensive perspective for interaction, enabling more accurate channel information aggregation. Incorporating $1\times1$ convolution introduces pixel-level channel interactions, further increasing mAP by 1.1\%, which suggests that local information is equally important for channel interactions.

\textbf{Global Channel Interaction Module Placement.} We use MLP as the baseline to examine the impact of placing the GCI module at different stages. The results in Table \ref{table5} suggest that placing the GCI module in the final two stages achieves the best performance, improving mAP by 1.6\% while reducing FLOPs and parameters by 32\% and 13\%, respectively. This can be attributed to the higher semantic level and richer channel information in the last two stages, making comprehensive channel interactions more beneficial.

\section{Conclusion}
In this paper, we propose the Sparse Mamba which achieves a superior balance between accuracy and efficiency for event-based object detection. The STCA module discards non-event and noisy tokens adaptively, reducing computational overhead significantly. The IPL-Scan and GCI modules capture the global context in spatial and channel dimensions, respectively. The IPL-Scan shortens the scan distance among high-information tokens and facilitates spatial contextual modeling. The GCI module aggregates channel information from the global spatial perspective, achieving global interaction within 3D space. Experimental results on three datasets demonstrate that our method achieves superior performance and computational efficiency.

\textbf{Limitation.} Our STCA relies on the assumption of temporal and spatial independence of event noise, which may limit its effectiveness in handling large bursts or clusters of noise. We will further consider integrating more effective semantic learning to better distinguish spatiotemporally continuous noisy tokens in future work.

\section{Acknowledgments}

This work is supported in part by the National Key Research and Development Program of China under Grant 2023YFC3081700, in part by the National Natural Science Foundation of China under Grants U24B20127 and 52172302.

\bibliography{aaai25}

\end{document}